\documentclass[10pt,twocolumn,letterpaper]{article}

\usepackage{iccv}              

\usepackage{wrapfig}
\usepackage{enumitem}
\usepackage{array}
\usepackage{booktabs}
\usepackage{tabularx}
\usepackage{array} 
\usepackage{graphicx}
\usepackage{multirow}
\usepackage{longtable}
\usepackage{pifont}
\usepackage[utf8]{inputenc} 
\usepackage[T1]{fontenc}    
\usepackage{url}            
\usepackage{booktabs}       
\usepackage{amsfonts}       
\usepackage{nicefrac}       
\usepackage{microtype}      
\usepackage{xcolor}         
\usepackage{amsmath}        
\usepackage[ruled,linesnumbered]{algorithm2e}  
\usepackage{graphicx}       
\usepackage{booktabs}       
\usepackage{colortbl}   
\usepackage{array}
\usepackage{booktabs}
\usepackage{tabularx}
\usepackage{array} 
\usepackage{graphicx}
\usepackage{multirow}
\usepackage{longtable}
\usepackage{pifont}
\definecolor{iccvblue}{rgb}{0.21,0.49,0.74}
\usepackage[pagebackref,breaklinks,colorlinks,allcolors=iccvblue]{hyperref}
\hypersetup{
    colorlinks=true,
    linkcolor=red,
    citecolor=blue,
    filecolor=magenta,      
    urlcolor=magenta,
    }
\def\paperID{3860} 
\def\confName{ICCV}
\def\confYear{2025}

\title{EA-ViT: Efficient Adaptation for Elastic Vision Transformer}

\author{
Chen Zhu$^{1,2}$\footnotemark[1]\quad
Wangbo Zhao$^{1}$\footnotemark[2]\quad
Huiwen Zhang$^{2}$\quad 
Samir Khaki$^{3}$\quad \\
Yuhao Zhou$^{1}$\quad 
Weidong Tang$^{1,2}$\quad 
Shuo Wang$^{4}$\quad
Zhihang Yuan$^{4}$\quad \\
Yuzhang Shang$^{5}$\quad
Xiaojiang Peng$^{6}$\quad
Kai Wang$^{1}$\quad
Dawei Yang$^{4}$\footnotemark[2]
\\
 $^{1}$National University of Singapore \quad
 $^{2}$Xidian University \quad
 $^{3}$University of Toronto \quad \\
 $^{4}$Houmo AI \quad 
 $^{5}$UCF \quad
 $^{6}$Shenzhen Technology University \quad
}

\begin{document}
\maketitle

\renewcommand{\thefootnote}{\fnsymbol{footnote}}
\footnotetext[1]{Work done during his undergraduate internship at NUS.}
\footnotetext[2]{Corresponding author.}

\begin{abstract}

Vision Transformers (ViTs) have emerged as a foundational model in computer vision, excelling in generalization and adaptation to downstream tasks. However, deploying ViTs to support diverse resource constraints typically requires retraining multiple, size-specific ViTs, which is both time-consuming and energy-intensive.
To address this issue, we propose an efficient ViT adaptation framework that enables a single adaptation process to generate multiple models of varying sizes for deployment on platforms with various resource constraints. 
Our approach comprises two stages. In the first stage, we enhance a pre-trained ViT with a nested elastic architecture that enables structural flexibility across MLP expansion ratio, number of attention heads, embedding dimension, and network depth. 
To preserve pre-trained knowledge and ensure stable adaptation, we adopt a curriculum-based training strategy that progressively increases elasticity. In the second stage, we design a lightweight router to select submodels according to computational budgets and downstream task demands. Initialized with Pareto-optimal configurations derived via a customized NSGA-II algorithm, the router is then jointly optimized with the backbone.
Extensive experiments on multiple benchmarks demonstrate the effectiveness and versatility of EA-ViT. The code is available at \url{https://github.com/zcxcf/EA-ViT}.

\end{abstract}

\section{Introduction}
\label{sec:intro}

\begin{figure}[t]
\label{fig:introduction}
\centering
\includegraphics[width=0.98\columnwidth]{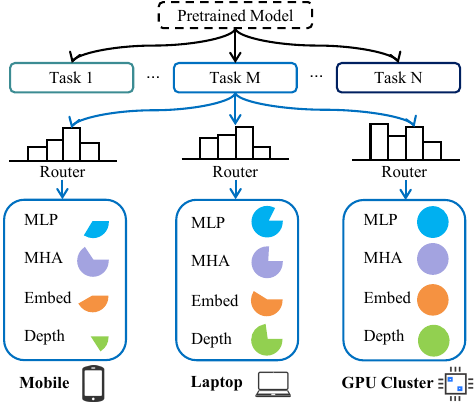} 
\vspace{-0.1in}
\caption{\textbf{Overview of the proposed EA-ViT framework.} Starting from a pre-trained ViT, EA-ViT generates task-specific elastic submodels through a router module. For each downstream task, the router selects a suitable submodel configuration—varying in MLP expansion ratio, number of attention heads, embedding dimension, and network depth. This ensures compatibility with the computational constraints of various target platforms, such as mobile devices, laptops, and GPU clusters.
}
\label{fig:introduction}
\end{figure}

Vision Transformers (ViTs)~\cite{vit,vitsurvey,VIT_survey,Deepvit,Maxvit} have significantly advanced the field of computer vision, delivering state-of-the-art results across a wide range of tasks~\cite{vit2,detr,vit3,vit4}. Pre-trained on large-scale datasets, ViTs exhibit strong generalization ability, enabling effective adaptation to diverse downstream applications.  Despite these advantages, deploying ViTs in real-world scenarios presents notable challenges, particularly in selecting the appropriate model size to suit diverse target platforms, ranging from resource-constrained mobile devices to high-performance GPU clusters.


A common approach to tackle this problem is to apply model pruning methods~\cite{prun1, prun2, prun3, prun4, prun5}, which reduce the model size to an appropriate level, followed by fine-tuning to restore performance for each downstream task. However, this strategy is inflexible  and presents two key challenges:  \emph{i)} Training separate models for each target device significantly increases computational cost and time requirement. \emph{ii)} Managing and storing multiple model variants complicates deployment and version control.

To support flexible development across different devices, recent works~\cite{onceforall,dynabert,matformer,hydravit} explore incorporating multiple submodels within a \textit{single} network. These methods allows for the selection of a model with an appropriate size, eliminating the need for additional training.
While promising, these methods still face notable limitations:
\textit{i)} These methods usually introduce elastic structures during the pre-training phase of ViTs, which requires additional fine-tuning for each submodel when transferred to downstream tasks.  \textit{ii)} More seriously, they typically support only a limited range of submodel sizes and rely on exhaustive search to find models that meet the requirements. As a results, \emph{efficiently deploying pre-trained ViTs across diverse devices and downstream tasks remains a pressing and unresolved challenge}.

 Motivated by these challenges, we propose an \textbf{E}fficient \textbf{A}daptation framework for \textbf{Vi}sion \textbf{T}ransformers (EA-ViT). As illustrated in Figure~\ref{fig:introduction}, our method applies elastic adaptation to a pre-trained ViT, producing a task-specific elastic variant tailored for downstream tasks. To support efficient deployment, we introduce a lightweight router that dynamically selects the optimal submodel configuration to meet the resource constraints of heterogeneous target platforms.

Our method is composed of two stages. In the first stage, we initially equip a pre-trained ViT with a nested elastic architecture. This design introduces structural flexibility along four key dimensions: MLP expansion ratio, number of attention heads, embedding dimension, and network depth. To effectively adapt the model while preserving its pre-trained knowledge, we employ a curriculum-based adaptation strategy that progressively incorporates smaller submodels, ensuring stable and efficient optimization.


In the second stage, we design a lightweight router to determine the submodel configuration across four dimensions based on task characteristics and computational constraints. To guide the router toward promising regions of the configuration space, we employ a customized NSGA-II algorithm to identify a set of candidate submodels that balance accuracy and computational cost. These configurations are used to initialize the router, which is then jointly optimized with the backbone to enable fine-grained exploration around the initial candidates. This method helps us to approach Pareto-optimal submodels and leverage limited training resources effectively to identify better-performing configurations.


Overall, EA-ViT introduces a unified adaptation framework that achieves an effective balance between accuracy and resource efficiency across various downstream tasks. We demonstrate the effectiveness of our approach on a wide range of benchmarks, including image classification, segmentation, and real-world datasets from domains such as medical imaging and remote sensing. Our main contributions are as follows:
\begin{itemize} 
    \item We introduce the first elastic adaptation framework for ViTs, enabling elasticity during the adaptation stage and significantly reducing both training and storage costs
    \item We propose a multi-dimensional elastic ViT design along with a lightweight router that selects optimal submodels based on device constraints and downstream task demands supported by an efficient training pipeline. 
    \item Extensive experiments on multiple datasets demonstrate that our method provides broader submodel size coverage and consistently achieves higher accuracy under varying constraints compared to existing elastic approaches.
\end{itemize}

\section{Related Work}
\begin{table}[t]
\centering
\caption{\textbf{Comparison between EA-ViT and previous approaches.} EA-ViT supports a broader range of elastic dimensions, resulting in a significantly larger submodel space. Taking ViT-Base as an example, our method generates up to $10^{12}$ times more submodels than the most flexible existing method.}
\label{tab:comparison}
\renewcommand{\arraystretch}{0.9}
\setlength{\tabcolsep}{4.2pt}
{\fontsize{9pt}{11pt}\selectfont
\begin{tabularx}{\linewidth}{lccccc}
\toprule
Method & MLP & MHA & Embed & Depth & Submodels \\
\midrule
DynaBERT~\cite{dynabert}   & \ding{51} & \ding{51} & \ding{55} & \ding{51} & $12$ \\
MatFormer~\cite{matformer} & \ding{51} & \ding{55} & \ding{55} & \ding{55} & \enspace $10^{7}$ \\
HydraViT~\cite{hydravit}   & \ding{55} & \ding{51} & \ding{51} & \ding{55} & $10$ \\
Flextron~\cite{flextron}   & \ding{51} & \ding{51} & \ding{55} & \ding{55} & \enspace\, $10^{14}$ \\
\midrule
\textbf{EA‑ViT (ours)}    & \ding{51} & \ding{51} & \ding{51} & \ding{51} & \enspace\, \textbf{$10^{26}$} \\
\bottomrule
\end{tabularx}
} 
\end{table}
\vspace{-0.1in}

\label{sec:related_work}
\paragraph{Network Architecture.}
Determining the optimal network architecture has long been a central goal in neural network research, irrespective of whether the networks are convolution-based (e.g., CNNs~\cite{resnet}) or attention-based (e.g., Transformers~\cite{vit}). Early efforts relied on manual design guided by empirical heuristics~\cite{densenet, alexnet}. Subsequently, Neural Architecture Search (NAS)~\cite{nas1,nas2} automated this design process by systematically exploring large architectural spaces, though at substantial computational cost. However, NAS-based approaches require re-searching and retraining models to satisfy the specific resource constraints whenever deploying to new platforms, limiting their practical scalability. 
To enhance deployment flexibility, recent approaches such as Once-for-All (OFA)~\cite{onceforall} and Slimmable Networks~\cite{slimmable} embed multiple submodels within a single elastic network, but remain mostly confined to convolutional architectures.

\paragraph{Elastic Transformer.}
Recently, elastic deployment strategies for Transformers~\cite{transformer} have gained increasing attention. DynaBERT~\cite{dynabert} leverages knowledge distillation to train width- and depth-adaptive models. MatFormer~\cite{matformer} adopts a Matryoshka-style design to control MLP expansion ratios, generating exponentially many submodels. HydraViT~\cite{hydravit} introduces elasticity in embedding size and attention heads, but only supports a limited number of submodels. Flextron~\cite{flextron} employs a router-based mechanism to build elastic MLP and MHA modules, yet relies on a surrogate model for training and focuses primarily on large language models.
In contrast, as shown in Table~\ref{tab:comparison}, our work targets Vision Transformers (ViTs)~\cite{vit} and introduces broader elasticity across four architectural dimensions: MLP expansion ratio, number of attention heads, embedding dimension, and network depth. More importantly, none of the aforementioned methods consider adapting a large, pre-trained ViT into an elastic structure during downstream task adaptation—a challenging yet crucial direction this work seeks to address.

\paragraph{Dynamic Network.}
Previous studies have explored dynamic adjustments of the computational graph conditioned on input samples to balance performance and efficiency. Typical approaches include dynamically varying the network depth~\cite{depth2,depth3} or width~\cite{width1,width2,width3, zhao2024dydit, zhao2025dydit++}, or selecting specialized subnetworks, such as mixture-of-experts (MoE)~\cite{moe,moe2,moe3} structures. In the context of Transformers, recent works have also introduced dynamic mechanisms along the token dimension~\cite{token2, zhao2024dynamic, zhao2025stitch}. For example, \cite{Mixture-of-depths} uses top-k token routing to dynamically allocate compute across sequence positions under a fixed budget and \cite{Mixture-of-nested-experts} leverages nested Mixture-of-Experts to prioritize tokens and adapt computation under varying budgets. However, these methods mainly adapt network structures according to input-dependent criteria. In contrast, our method explicitly selects submodel configurations based on computational constraints and downstream task characteristics.

\begin{algorithm*}[t]
\small
\caption{Curriculum Elastic Adaptation}
\label{pseudocode}
\KwIn{%
  Pre-trained ViT backbone with $L$ blocks : $\mathcal{M}_0$; \quad
  Total training steps : $T$; \quad
  Expansion step set : $\mathcal{S}_{\mathrm{exp}} \subset \{1,\dots,T\}$; \quad
  Initial upper bounds: $R_{\max},\; H_{\max},\; E_{\max},\; n_{\max}$; \quad
  Per-expansion changes: $\Delta_R,\; \Delta_H,\; \Delta_E,\; \Delta_n$;
}
\KwOut{Adapted elastic ViT $\mathcal{M}_{\mathrm{elastic}}$}

\BlankLine
\textbf{// Step 0: Initialize sampling ranges (start with largest model)}\\
$R_{\min} \gets R_{\max};\quad
 H_{\min} \gets H_{\max};\quad
 E_{\min} \gets E_{\max};\quad
 n_{\max} \gets 0$\;

\BlankLine
\For{$t \gets 1$ \KwTo $T$}{
  \textbf{// Step 1: Expand elasticity at scheduled steps}\\
  \If{$t \in \mathcal{S}_{\mathrm{exp}}$}{
    $R_{\min} \gets R_{\min} - \Delta_R$;\quad
    $H_{\min} \gets H_{\min} - \Delta_H$;\quad
    $E_{\min} \gets E_{\min} - \Delta_E$;\quad
    $n_{\max} \gets n_{\max} + \Delta_n$\quad
  }

  \BlankLine
  \textbf{// Step 2: Sample submodel hyperparameters}\\
  $R \sim \mathcal{U}(R_{\min}, R_{\max})$;\quad
  $H \sim \mathcal{U}(H_{\min}, H_{\max})$;\quad
  $E \sim \mathcal{U}(E_{\min}, E_{\max})$;\quad
  $s \sim \mathcal{U}\{0, \dots, n_{\max}\}$ \\
Select $S \subseteq \{1,\dots,L\}$ with $|S| = s$; for $l=1$ to $L$: $D^{(l)} \gets 0$ if $l \in S$, else $1$;\;

  \BlankLine
  \textbf{// Step 3: Build and train sampled submodel}\\
  $\mathcal{M}_t \gets \textsc{BuildSubModel}(\mathcal{M}_{t-1}, R, H, E, \{D^{(l)}\})$;\
  $\textit{loss} \gets \mathcal{L}(\mathcal{M}_t(\text{data}))$;
  Backpropagate $\nabla_\theta \textit{loss}$ and update weights;
}
\Return $\mathcal{M}_{\mathrm{elastic}} \gets \mathcal{M}_T$
\end{algorithm*}

\section{Method}
\label{sec:method}

\begin{figure}[t]
\centering
\includegraphics[width=\columnwidth]{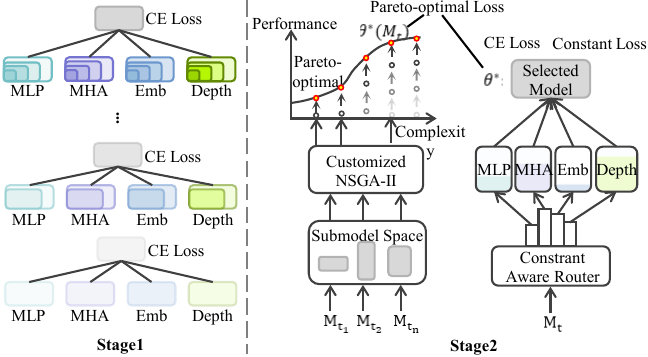} 
\caption{\textbf{Overall pipeline of our method.} In Stage 1, we construct a Multi-Dimensional Elastic Architecture and introduce elasticity into a pre-trained ViT through Curriculum-based Elastic Adaptation. In Stage 2, Pareto-optimal Submodel Search is employed to identify promising submodel configurations used to initialize the router. Then, the router and backbone are jointly optimized to further enhance submodel performance.}
\label{fig:method}
\end{figure}

As illustrated in Figure~\ref{fig:method}, the training procedure of EA-ViT comprises two stages. In Stage 1, we first construct a Multi-Dimensional Elastic Architecture (Section~\ref{Multi-Dimensional Elastic Architecture}) and then apply a Curriculum-based Elastic Adaptation strategy (Section ~\ref{Curriculum Elastic Adaptation}) for preliminary training of the elastic structure. In Stage 2, we employ a customized NSGA-II algorithm to initialize the router (Section~\ref{Pareto-Optimal Submodel Search}) and subsequently train the router jointly with the backbone network (Section~\ref{Constraint-Aware Router}).



\subsection{Multi-Dimensional Elastic Architecture}
\label{Multi-Dimensional Elastic Architecture}
Each block in a ViT consists of two core components: a Multi-Head Attention (MHA) module and a Multi-Layer Perceptron (MLP) module, both connected via residual pathways and followed by Layer Normalization. The primary architectural factors affecting model capacity and computational cost are the MLP expansion ratio, number of attention heads, embedding dimension, and network depth.

Leveraging these dimensions, we construct an elastic ViT architecture that supports flexible configuration along all four axes. The embedding dimension is fixed across all layers, while the number of attention heads and MLP expansion ratios can vary per layer. The depth dimension controls whether we apply residual connections or skip blocks via identity mappings. These dimensions together define a large architectural search space, denoted as $\theta \in \Theta$, where
\begin{equation}
\theta = \Bigl(\{R^{(l)}\}_{l=1}^L, \{H^{(l)}\}_{l=1}^L, E, \{D_{\text{MLP}}^{(l)}, D_{\text{MHA}}^{(l)}\}_{l=1}^L\Bigr).
\label{equ1}
\end{equation}
Here, the components are defined as follows:
\begin{itemize}
    \item \(L\) denotes the total number of network \textbf{L}ayers.
    \item \(R^{(l)}\) indicates the expansion \textbf{R}atio of the hidden dimension in the MLP at the \(l\)-th layer.
    \item \(H^{(l)}\) represents the number of active attention \textbf{H}eads at the \(l\)-th layer.
    \item \(E\) denotes the unified \textbf{E}mbedding dimension shared across all layers.
    \item \(D_{\text{MLP}}^{(l)}\) and \(D_{\text{MHA}}^{(l)}\) are binary indicators denoting whether the MLP and MHA blocks are activated at the \(l\)-th layer, respectively, thereby adjusting the network \textbf{D}epth.
\end{itemize}

To promote parameter sharing across submodels, we adopt a nested elastic structure. Embedding dimensions, MLP hidden units, and attention heads are ranked by estimated importance, allowing more critical components to be shared across a broader range of submodels. Additional details on the ranking methodology are provided in the supplementary material \ref{Importance Rearrangement}.

\subsection{Curriculum Elastic Adaptation}
\label{Curriculum Elastic Adaptation}
Training numerous submodels simultaneously can lead to parameter interference due to conflicting gradient signals. To mitigate this, we employ a phased curriculum learning strategy that gradually expands model elasticity. Specifically, training starts with minimal elasticity — using the original pre-trained model — and progressively introduces a broader range of submodels, including smaller variants. This controlled expansion follows curriculum learning principles, moving from simpler to more complex configurations.



At the initial stage, we train only the largest submodel. As training progresses, we incrementally expand the sampling ranges for the architectural parameters defined in Equation~\ref{equ1}. Specifically, the MLP expansion ratio \(R\), the number of attention heads \(H\), and the embedding dimension \(E\) are each sampled from uniform distributions with gradually decreasing lower bounds as a function of the training step:
\begin{equation}
\resizebox{\columnwidth}{!}{$
R\sim\mathcal{U}(R_{\min}^{(t)},R_{\max}),\;H\sim\mathcal{U}(H_{\min}^{(t)},H_{\max}),\;E\sim\mathcal{U}(E_{\min}^{(t)},E_{\max})
$}, 
\end{equation}

where the lower bounds \(R_{\min}^{(t)}, H_{\min}^{(t)}, E_{\min}^{(t)}\) are updated at scheduled training steps to progressively enlarge the submodel space.
For the depth parameter \(D\), we adopt a stochastic sampling strategy to control the number of skipped layers. Initially, all layers are active, i.e., \(D^{(l)} = 1\) for all \(l = 1, \dots, L\). As training proceeds, we gradually increase the number of skipped layers. Specifically, for a given number \(n\), we randomly select a subset \(S \subseteq \{1, \dots, L\}\) with \(|S| = n\), and set \(D^{(l)} = 0\) for all \(l \in S\), indicating the corresponding layers are skipped.
The full procedure is presented in the pseudocode shown in Algorithm~\ref{pseudocode}.

\begin{figure*}[t]
\centering
\includegraphics[width=\textwidth]{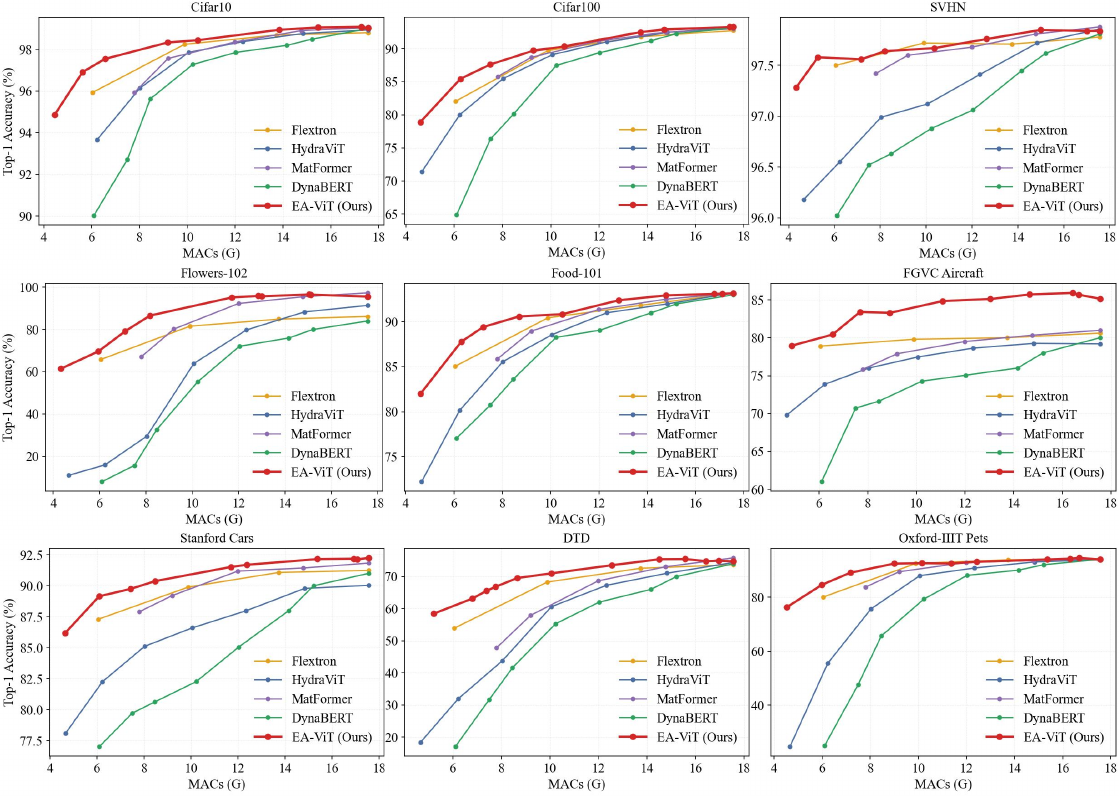} 
\caption{\textbf{MACs(G) vs. Top-1 accuracy curves of our method and prior approaches across nine image classification benchmarks.} Our method consistently outperforms existing methods across nearly all MACs constraints on all datasets.}
\label{fig:9results}
\end{figure*}



\begin{table*}[htbp]
  \centering
  \scriptsize                        
  \setlength{\tabcolsep}{3pt}        
  \renewcommand{\arraystretch}{1.05} 
  \caption{\textbf{Top‐1 accuracy across nine classification benchmarks.} We adopt 8 GMACs constraint. The best scores are in \textbf{bold}.}
  \label{tab:9result}

  \resizebox{\textwidth}{!}{%
    \begin{tabular}{@{} l *{9}{c} @{}}
      \toprule
      Method          & Cifar10~\cite{cifar100} & Cifar100~\cite{cifar100} & SVHN~\cite{svhn} & Flowers~\cite{flower102}
                      & Food101~\cite{food101} & Aircraft~\cite{fgvc} & Cars~\cite{car}
                      & DTD~\cite{DTD} & Pets~\cite{pet} \\
      \midrule
      DynaBERT~\cite{dynabert}   & 94.24 & 78.30 & 96.58 & 24.53 & 82.24 & 71.19 & 80.19 & 36.88 & 57.05 \\
      MatFormer~\cite{matformer} & 96.17 & 86.18 & 97.45 & 69.00 & 86.31 & 76.13 & 88.10 & 49.25 & 84.53 \\
      HydraViT~\cite{hydravit}   & 96.11 & 85.39 & 96.98 & 29.27 & 85.49 & 75.95 & 85.07 & 43.53 & 75.32 \\
      Flextron~\cite{flextron}   & 97.11 & 85.95 & 97.61 & 73.80 & 87.79 & 79.36 & 88.62 & 61.21 & 86.35 \\
      \rowcolor{gray!10}
      EA-ViT (ours)              & \textbf{97.98} & \textbf{88.20} & \textbf{97.63} & \textbf{85.39}
                                & \textbf{90.05} & \textbf{83.37} & \textbf{90.09} & \textbf{67.56} & \textbf{90.57} \\
      \bottomrule
    \end{tabular}%
  }
\end{table*}

\


\begin{figure*}[t]
\centering
\includegraphics[width=\textwidth]{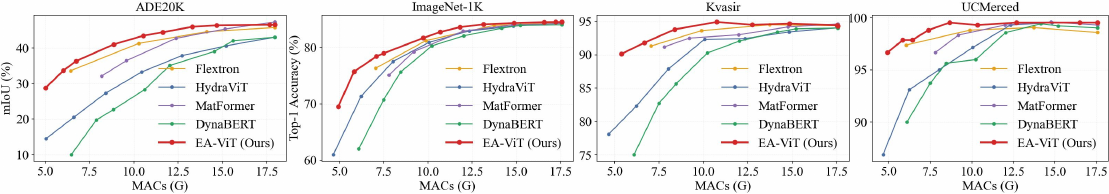} 
\caption{\textbf{Comparison of MACs vs. mIoU/Top-1 accuracy between our method and prior approaches on four more challenging datasets:}  Semantic segmentation benchmark (ADE20K \cite{ade20k}), a large-scale image classification dataset (ImageNet-1K \cite{imagenet}), a medical image classification dataset (Kvasir \cite{kvasir}), and a remote sensing classification dataset (UCMerced \cite{ucm}). Our method consistently outperforms existing approaches across nearly all MACs constraints on these more difficult tasks.}
\label{fig:seg and medical and remote}
\end{figure*}

\subsection{Pareto-Optimal Submodel Search}
\label{Pareto-Optimal Submodel Search}
To accelerate router convergence within the exponentially large submodel space, we first identify a set of promising candidate submodels for initialization. We leverage the NSGA-II evolutionary algorithm~\cite{nsga} as it is inherently well-suited for this task: its crossover operation is akin to the structured recombination of architectural components across multiple dimensions, while its mutation process promotes diversity and prevents premature convergence, enhancing the robustness of submodel discovery.


However, directly applying standard NSGA-II can introduce several challenges in our setting, including redundancy among selected individuals and insufficient coverage across different model complexity levels. To mitigate these issues, we introduce a partitioned selection strategy combined with an iterative crowding-distance-based selection mechanism. Specifically, we partition the normalized complexity space and perform independent selection within each partition, while enforcing a minimum complexity gap between selected individuals. This approach significantly improves population diversity and ensures broad and balanced coverage of the submodel search space. Further implementation details are provided in the supplementary material \ref{nsga}.

\subsection{Constraint-Aware Router}
\label{Constraint-Aware Router}
Deploying ViTs in real-world settings requires selecting submodel configurations that align with the resource constraints of the target device.
We introduce a constraint-aware router that predicts the optimal submodel configuration given a specific resource budget, such as the number of Multiply-Accumulate operations (MACs).
While MACs are the primary constraint considered in this work, the framework is easily extensible to other metrics such as parameter count.
\paragraph{Construction.}

We implement the router as a lightweight two-layer MLP that maps a normalized target constraint \(M_t \in [0,1]\) to a corresponding submodel configuration \(\theta\). Given the normalized MACs budget as input, the router outputs the architectural parameters defined in Equation~\ref{equ1}:
\begin{equation}
\resizebox{\columnwidth}{!}{$
\theta = \Bigl(\{R^{(l)}\}_{l=1}^L, \{H^{(l)}\}_{l=1}^L, E, \{D_{\text{MLP}}^{(l)}, D_{\text{MHA}}^{(l)}\}_{l=1}^L\Bigr)
= \mathrm{Router}(M_t).
$}
\label{equ3}
\end{equation}

Since the router's output \(\theta\) involves discrete architectural decisions, we adopt the Gumbel-Sigmoid~\cite{gumbelsigmoid} trick to enable gradient-based optimization. Specifically, the router's logits are perturbed with Gumbel noise and passed through a temperature-controlled sigmoid to produce a continuous relaxation:
\begin{equation}
\begin{aligned}
y_{\text{soft}} &= \operatorname{Sigmoid}\left(\frac{\text{logits} + G_1 - G_2}{\tau}\right),
y_{\text{hard}} = \mathbb{I}(y_{\text{soft}} > \delta),
\end{aligned}
\end{equation}
where \(G_1, G_2 \sim \text{Gumbel}(0,1)\) are independent noise samples, \(\tau\) is a temperature parameter controlling the approximation sharpness and \(y_{\text{hard}}\) is obtained by thresholding \(y_{\text{soft}}\).


To preserve gradient flow during training, we use the straight-through estimator:
\begin{equation}
y = y_{\text{hard}} - \text{detach}(y_{\text{soft}}) + y_{\text{soft}},
\end{equation}
where $\text{detach(.)}$ denotes removing gradients during backpropagation. This formulation allows the router to make discrete submodel selections while remaining fully differentiable for end-to-end optimization.

\paragraph{Training.}
To jointly optimize submodel performance and ensure adherence to computational constraints, we formulate a bi-objective minimization problem that balances predictive accuracy with resource usage as follows:
\begin{equation}
\min_{\theta}
\left(\mathcal{L}_{\mathrm{CE}}(\theta), \;\; \left| \frac{\mathrm{MACs}(\theta)}{M_0} - M_t \right| \right),
\end{equation}
where \(M_t \in [0, 1]\) denotes the normalized target MACs, and \(M_0\) is the MACs of the largest model in the design space.

Given a sequence length \( N \) and the dimension of each attention head \( d_{\mathrm{head}} \), the MACs for a given submodel \(\theta\) can be computed as:
\begin{equation}
\begin{aligned}
\mathrm{MACs}(\theta)
&= \sum_{l=1}^L \Bigl[
    D_{\mathrm{MLP}}^{(l)} \cdot 2NE^2 R^{(l)} \\
&\quad\;+\; D_{\mathrm{MHA}}^{(l)} \cdot N d_{\mathrm{head}} H^{(l)} \bigl(4E + 2N\bigr)
\Bigr].
\end{aligned}
\end{equation}

The first term captures the MLP complexity, while the second term accounts for the computational cost of MHA, including both projection and attention operations.

Let \(\theta^*(M_t)\) denote the submodel on the Pareto front discovered by Pareto-Optimal Submodel Search whose MACs most closely match the target constraint \(M_t\). The unified loss function used in Stage 2 is formulated as:
\begin{equation}
\mathcal{L}
= \mathcal{L}_{\mathrm{CE}}
+ \lambda_1 \left( \frac{\mathrm{MACs}(\theta)}{M_0} - M_t \right)^2
+ \lambda_2 \left\| \theta - \theta^*(M_t) \right\|_2^2,
\end{equation}
where \(\lambda_1\) penalizes deviation from the computational constraint, and \(\lambda_2\) controls the strength of early-stage supervision based on the reference configuration \(\theta^*(M_t)\).


In the early phase of Stage~2, we set a relatively large $\lambda_2$ to guide the router toward known Pareto-optimal submodels, thereby accelerating convergence. As training progresses, $\lambda_2$ is gradually annealed to zero, allowing the router to autonomously explore the search space and potentially discover submodel configurations that surpass those in the initial Pareto front.

\section{Evaluation}

\subsection{Experiment Setup}

\paragraph{Compared Methods.}
We compare our method against several state-of-the-art approaches that incorporate elastic structures into Transformer architectures. Specifically, we evaluate against DynaBERT~\cite{dynabert}, which introduces elastic width and depth via knowledge distillation; MatFormer~\cite{matformer}, one of the earliest works to explore elasticity in Transformers; HydraViT~\cite{hydravit}, which builds elasticity based on varying the number of attention heads; and the recently proposed Flextron~\cite{flextron}, a powerful framework that integrates elasticity into both the MLP and MHA modules and employs a router for submodel selection.
We integrate these methods into the adaptation process to incorporate elastic structures.

\paragraph{Datasets.}
To comprehensively evaluate the effectiveness and generalizability of our method, we conduct experiments across a wide range of downstream tasks. Specifically, we perform adaptation on nine image classification datasets: Cifar10~\cite{cifar100}, Cifar100~\cite{cifar100}, SVHN~\cite{svhn}, Flowers-102~\cite{flower102}, Food-101~\cite{food101}, FGVC Aircraft~\cite{fgvc}, Stanford Cars~\cite{car}, DTD~\cite{DTD}, and Oxford-IIIT Pets~\cite{pet}.
To assess the applicability of our approach to dense prediction tasks, we further evaluate it on one widely-used semantic segmentation benchmarks: ADE20K~\cite{ade20k}. We also evaluate our method on the difficult ImageNet-1K~\cite{imagenet} datasets.
In addition, to demonstrate the robustness of our method in real-world scenarios, we conduct experiments on two real-world application datasets: Kvasir~\cite{kvasir} for medical image analysis, and UCMerced~\cite{ucm} for remote sensing applications.

\begin{table}[t]
  \centering
  \scriptsize
  \setlength{\tabcolsep}{3pt}
  \renewcommand{\arraystretch}{1.0}
  \setlength{\tabcolsep}{1.1pt}
  \caption{\textbf{The impact of progressively enabling elasticity across four architectural dimensions:} MLP, MHA, Embed, and Depth. We report the minimum achievable MACs, the highest Top-1 accuracy, and the area under the MACs--accuracy curve (AUC) along with its slope. As more dimensions are made elastic, the architecture achieves better accuracy and broader MACs coverage. Experiments are conducted on Cifar100~\cite{cifar100}.}
  \label{tab:ablation_dim}
  {\fontsize{8.0pt}{11pt}\selectfont
  \begin{tabularx}{\columnwidth}{@{} cccc | cccc @{}}
    \toprule
    MLP & MHA & Embed
    & Depth
        & MACs$_{\min}$ ($\downarrow$) & Acc$_{\max}$ ($\uparrow$) & Slope ($\downarrow$) & AUC ($\uparrow$) \\
    \midrule
    \checkmark & -- & -- & --          & 7.80 & 93.01 & 1.658 & 0.505 \\
    \checkmark & \checkmark & -- & --    & 6.04 & 93.12 & 1.515 & 0.646 \\
    \checkmark & \checkmark & \checkmark & -- & 5.11 & 93.31 & 1.372 & 0.651 \\
    \checkmark & \checkmark & \checkmark & \checkmark
               & \textbf{4.60} & \textbf{93.32} & \textbf{1.235} & \textbf{0.663} \\
    \bottomrule
  \end{tabularx}
  }
\end{table}

\paragraph{Implementation Details.}
All experiments are conducted on NVIDIA A100 GPUs. For adaptation and expansion, we adopt ViT-Base as the backbone, initialized with weights pre-trained on ImageNet-21K~\cite{ini21k}.
During the Curriculum Elastic Adaptation stage, we first trained the model to incorporate elasticity. We then trained the router independently for 1,000 steps, guiding it to output submodels that match the Pareto-optimal configurations identified in the previous search stage. Finally, the router and ViT backbone were jointly optimized. All training phases employed a cosine decay learning rate schedule with linear warm-up.
For a fair comparison, all baseline models and our ViT backbone were trained for the same number of epochs on each downstream dataset. Additional training details, including the specific learning rates and number of epochs for each dataset, are provided in the supplementary materials~\ref{Training Hyperparameter Settings}.

\subsection{Comparison with State-of-the-Art Methods}
As shown in Figure~\ref{fig:9results}, we conduct a comprehensive comparison between our method and prior state-of-the-art approaches across nine image classification benchmarks. The accuracy–MACs curves clearly demonstrate that our method consistently outperforms existing methods across all datasets and under nearly all computational budgets, with particularly significant advantages in the low-MACs regime.
Moreover, we observe that different datasets exhibit varying sensitivities to model size. For simpler datasets such as SVHN~\cite{svhn}, accuracy degrades slowly as MACs decrease, indicating a lower dependence on model capacity. In contrast, more complex datasets such as DTD~\cite{DTD} and Flowers-102~\cite{flower102} experience a steep accuracy drop under constrained MACs, highlighting their greater need for model expressiveness. These observations suggest that our method can adaptively select appropriate submodel sizes based on dataset characteristics, offering a practical advantage in real-world deployment scenarios.

As shown in Table~\ref{tab:9result}, we report the Top-1 accuracy of various methods under an 8 GMACs computational budget. Our method consistently achieves substantial improvements over existing baselines. For instance, on the Flowers-102~\cite{flower102} dataset, our approach outperforms the best prior method by a remarkable margin of 26.39\%, demonstrating its strong ability to handle complex visual patterns under limited computational budgets. Comparision under other MACs constraint can be seen in the supplementary material~\ref{More Detailed Results}.

We further evaluate our method on more challenging tasks, including image segmentation, large-scale image classification, medical image classification, and remote sensing image classification. As illustrated in Figure~\ref{fig:seg and medical and remote}, our method consistently outperforms previous approaches across all MACs configurations, demonstrating strong generalization capabilities in complex scenarios. These results highlight the versatility and practical value of our approach across diverse real-world applications.

\subsection{Ablation Study}
To assess the effectiveness of each component in our proposed method, we conduct comprehensive ablation studies.

\begin{figure}[t]
\centering
\includegraphics[width=1.0\columnwidth]{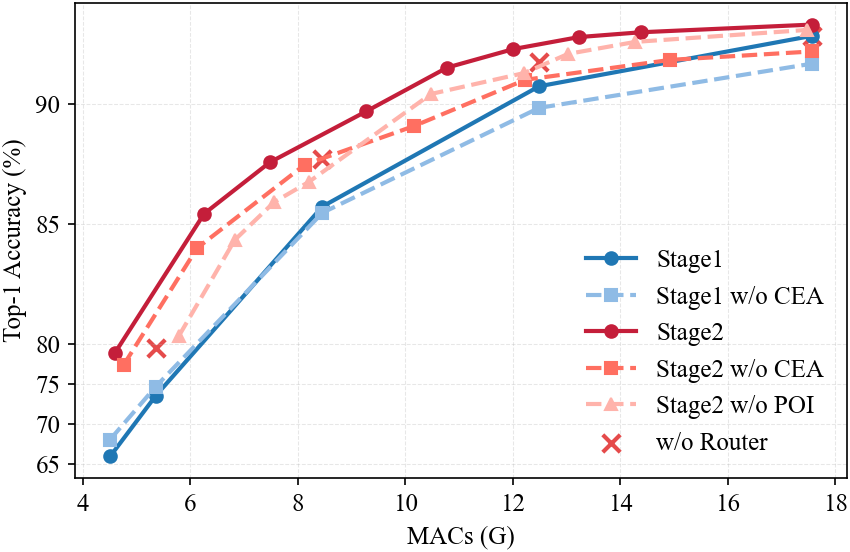} 
\vspace{-0.1in}
\caption{\textbf{The contributions of key components in our framework:} The router, Curriculum Elastic Adaptation (CEA) and Pareto-Optimal Initialization (POI). Experiments are conducted on Cifar100~\cite{cifar100}.}
\vspace{-0.2in}
\label{fig:ablation module}
\end{figure}

\paragraph{Multi-Dimensional Elastic Architecture.}
We investigate whether increasing the number of controllable architectural dimensions leads to improved model performance under varying computational budgets. To this end, we compute four key metrics from the accuracy–MACs curves corresponding to different combinations of elastic dimensions: the minimum achievable MACs, the highest Top-1 accuracy, the average slope, and the Area Under the Curve (AUC).
As summarized in Table~\ref{tab:ablation_dim}, utilizing only the MLP expansion ratio leads to limited MACs coverage and notable performance degradation, due to the coarse granularity of control along a single architectural dimension. This is reflected in the relatively high slope observed.
In contrast, progressively incorporating additional dimensions—such as the number of attention heads, embedding size, and layer depth—consistently increases MACs coverage, improves the maximum achievable accuracy, enhances the AUC, and reduces the slope. These results underscore the importance of multi-dimensional elasticity for robust and efficient submodel deployment across a wide range of computational constraints.

\paragraph{Effectiveness of the Router.}
As shown in Figure~\ref{fig:ablation module}, we conduct an ablation study to assess the impact of the router. We compare submodels manually selected with fixed architectures against those automatically selected by the router, ensuring all models are trained for the same number of epochs. The manually selected submodels consistently lie below the MACs–accuracy curve achieved by the router, indicating inferior performance under comparable computational budgets. These results validate the effectiveness of our routing mechanism in identifying high-quality, resource-efficient submodels.
\paragraph{Curriculum Elastic Adaptation.}
We conduct an ablation study to evaluate the effectiveness of Curriculum Elastic Adaptation (CEA) during Stage 1 training, as shown in Figure~\ref{fig:ablation module}. Removing CEA results in the simultaneous training of all submodels from the beginning, which leads to a disproportionately high sampling frequency of smaller submodels. While this slightly improves performance under low-MACs conditions, it negatively impacts the optimization of larger submodels, resulting in noticeable performance degradation at higher computational budgets—likely due to interference in shared parameter learning.

After Stage 2 refinement, models trained with CEA consistently outperform those without it across the entire MACs range. These results underscore the importance of curriculum-based scheduling in promoting training stability and preserving the performance of high-capacity submodels.
\paragraph{Pareto-Optimal Initialization.}
We conduct an ablation study to assess the impact of Pareto-Optimal Initialization (POI) on the router’s performance during Stage 2, as shown in Figure~\ref{fig:ablation module}. Instead of initializing the router with randomly selected submodels, POI leverages architectures sampled from the Pareto front. This informed initialization allows the router to focus its search on a promising region of the architectural space from the outset. By prioritizing submodels with high potential, POI helps the training process make more effective use of limited computational resources. As a result, the router can more efficiently identify submodels that are both performant and resource-efficient, ultimately leading to improved overall system performance.

\subsection{Analysis}
\begin{figure}[t]
\centering
\includegraphics[width=0.99\columnwidth]{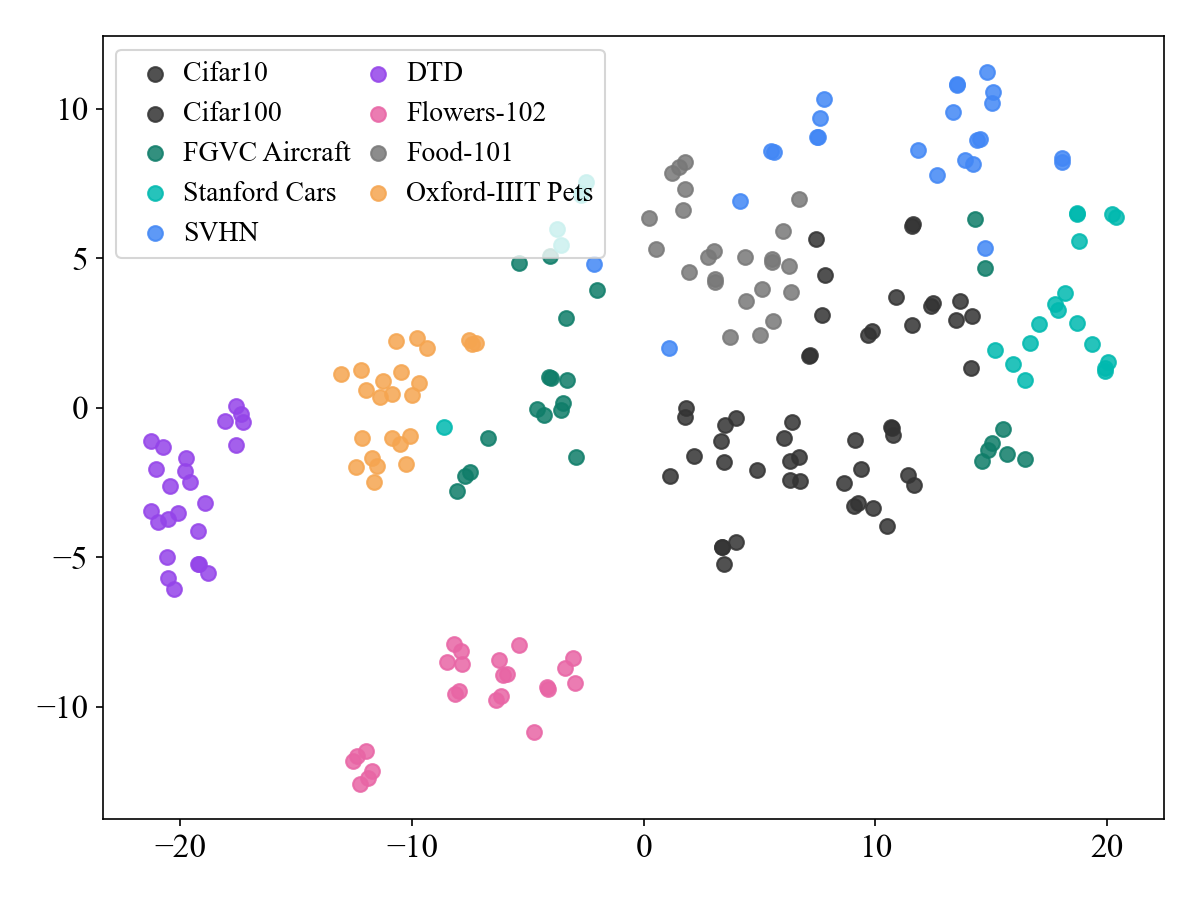} 
\vspace{-0.1in}
\caption{\textbf{t-SNE visualization of submodel structures selected by the router.} Experiments are conducted within a normalized MACs range of 0.4 to 0.6 across nine datasets. Submodels from similar datasets tend to cluster closely, while those from dissimilar datasets exhibit distinct structural differences.}
\vspace{-0.2in}
\label{fig:TSNE}
\end{figure}

We employ t-SNE to visualize the distribution of submodel architectures obtained across nine image classification datasets. As shown in Figure \ref{fig:TSNE}, we focus on submodels with MACs in the range of 0.4 to 0.6, and observe that different downstream tasks exhibit distinct structural preferences. Furthermore, the degree of divergence in submodel choices varies significantly across datasets.

Specifically, datasets such as DTD~\cite{DTD} (oriented toward texture recognition) and SVHN~\cite{svhn} (digit classification) demonstrate submodel preferences that deviate markedly from those of general object recognition datasets. This is reflected by their embeddings forming distant outliers in the t-SNE space. In contrast, structurally similar datasets—such as Cifar10~\cite{cifar100} and Cifar100~\cite{cifar100}, or FGVC-Aircraft~\cite{fgvc} and Stanford Cars~\cite{car}—produce submodels with highly similar configurations, as indicated by the tight clustering of their points. These observations underscore the importance of dataset-specific adaptability in the design of elastic architectures. They further demonstrate the effectiveness of our router in identifying optimal submodel configurations under varying computational constraints across diverse datasets. Detailed architectural specifications for each dataset are provided in the supplementary material~\ref{Visualization of Submodel Architectures}.

\section{Conclusion}

In this paper, we introduce the first elastic adaptation framework for Vision Transformers, enabling architectural elasticity during the adaptation stage and substantially reducing both training and deployment costs.
Our approach features a multi-dimensional elastic ViT design, coupled with a lightweight router that dynamically selects optimal submodels based on different constraints. We incorporate a curriculum learning strategy that gradually expands architectural elasticity during adaptation to ensure stable optimization. Furthermore, we accelerate router convergence by initializing it with Pareto-optimal submodels identified via a customized NSGA-II algorithm, followed by joint training with the backbone.
Extensive experiments across a wide range of benchmarks demonstrate that our method consistently outperforms existing approaches and offers new insights into the structural preferences of ViTs under varying resource budgets and downstream task demands. We hope this work will inspire future research on efficient, flexible, and scalable ViT adaptation for real-world deployment.

\paragraph{Acknowledgments.}
This work was partially supported by the National Research Foundation, Singapore under its AI Singapore Programme (AISG Award No: AISG2-PhD-2021-08-008), the National Natural Science Foundation of China (62176165), the Stable Support Projects for Shenzhen Higher Education Institutions (20220718110918001), the Scientific Research Capacity Enhancement Program for Key Construction Disciplines in Guangdong Province under Grant 2024ZDJS063. Yang You’s research group is being sponsored by NUS startup grant (Presidential Young Professorship), Singapore MOE Tier-1 grant, ByteDance grant, ARCTIC grant, SMI grant (WBS number: A8001104-00-00), Alibaba grant, and Google grant for TPU usage.

\small
\bibliographystyle{ieeenat_fullname}
\bibliography{main}

\clearpage

\appendix

In this Supplementary Material, we provide additional details of our method. Section \ref{method details} elaborates on key implementation aspects, including the importance-based rearrangement strategy and the customized improvements to the NSGA-II algorithm. Section \ref{Training Hyperparameter Settings} describes the training setup for Curriculum Elastic Adaptation and outlines the hyperparameters used for different datasets. Section \ref{More Detailed Results} presents additional results, including comparisons under more evaluation metrics and detailed accuracy across multiple MACs levels. Section \ref{Visualization of Submodel Architectures} presents the architectural configurations of the selected submodels under varying MACs constraints across multiple datasets. Section \ref{Potential Applications} discusses some potential applications.

\section{Method Details}
\label{method details}
\subsection{Importance Rearrangement}
\label{Importance Rearrangement}

Before constructing the elastic structure, we perform importance-based rearrangement across the expandable dimensions in the Vision Transformer (ViT). 
This process prioritizes critical units, ensuring they are shared across a larger number of submodels within the nested elastic structure. 
As a result, important units are sampled more frequently during training, receive more updates, and achieve better robustness and transferability across different submodel configurations.

\paragraph{Embedding Dimension Importance.}
A Vision Transformer (ViT) consists of a stack of \(L\) identical Transformer blocks, where each block includes a Multi-Head Attention (MHA) module and a Feed-Forward Network (MLP), both equipped with residual connections and Layer Normalization (LN).

Given the initial input sequence \(\mathbf{x}^{(0)} \in \mathbb{R}^{N \times D_{\text{emb}}}\), the feature propagation across blocks is:
\begin{equation}
\mathbf{x}^{(l+1)} = \text{Block}^{(l)}(\mathbf{x}^{(l)}), \quad l = 0, \dots, L-1.
\end{equation}
Each block computes:
\begin{equation}
\begin{aligned}
\mathbf{y}^{(l)} &= \mathbf{x}^{(l)} + \text{MHA}(\text{LN}(\mathbf{x}^{(l)})), \\
\mathbf{x}^{(l+1)} &= \mathbf{y}^{(l)} + \text{MLP}(\text{LN}(\mathbf{y}^{(l)})).
\end{aligned}
\end{equation}

To assess the importance of each embedding channel, we use the final block output \(\mathbf{x}^{(L)}\). For each embedding dimension \(i \in \{1, \dots, D_{\text{emb}}\}\), the importance score is computed as:
\begin{equation}
F^{(i)}_{\text{emb}} = \sum_{\mathbf{s} \in \mathcal{X}} \left\| \mathbf{x}^{(L,i)}(\mathbf{s}) \right\|_1,
\end{equation}
where \(\mathcal{X}\) denotes the sampled input set, and \(\mathbf{x}^{(L,i)}(\mathbf{s})\) represents the activation of the \(i\)-th embedding dimension for sample \(\mathbf{s}\).

\paragraph{MLP Hidden Dimension Importance.}
Each MLP module consists of two linear layers with an intermediate hidden dimension \(D_{\text{hid}} = r \times D_{\text{emb}}\), where \(r\) is the expansion ratio. 
Given an input \(\mathbf{z} \in \mathbb{R}^{N \times D_{\text{emb}}}\), the MLP computation proceeds as:
\begin{equation}
\mathbf{h} = \mathbf{z} \mathbf{W}_1, \quad \mathbf{z}' = \phi(\mathbf{h}) \mathbf{W}_2,
\end{equation}
where \(\mathbf{W}_1 \in \mathbb{R}^{D_{\text{emb}} \times D_{\text{hid}}}\), \(\mathbf{W}_2 \in \mathbb{R}^{D_{\text{hid}} \times D_{\text{emb}}}\), and \(\phi(\cdot)\) is the activation function (e.g., GELU).

The importance of each hidden neuron \(j \in \{1, \dots, D_{\text{hid}}\}\) is measured as:
\begin{equation}
F^{(j)}_{\text{mlp}} = \sum_{\mathbf{s} \in \mathcal{X}} \left\| \phi\left( \mathbf{h}^{(l,j)}(\mathbf{s}) \right) \right\|_1,
\end{equation}
where \(\mathbf{h}^{(l,j)}(\mathbf{s})\) denotes the activation of the \(j\)-th hidden neuron at layer \(l\) for sample \(\mathbf{s}\).

\paragraph{MHA Attention Head Importance.}
For each block, the MHA module applies self-attention to the input \(\mathbf{z}^{(l)}\). 
For head \(m \in \{1, \dots, h^{(l)}\}\), the queries, keys, and values are computed as:
\begin{equation}
\mathbf{q}^{(m)} = \mathbf{z}^{(l)} \mathbf{W}_Q^{(m)}, \quad 
\mathbf{k}^{(m)} = \mathbf{z}^{(l)} \mathbf{W}_K^{(m)}, \quad 
\mathbf{v}^{(m)} = \mathbf{z}^{(l)} \mathbf{W}_V^{(m)},
\end{equation}
where \(\mathbf{W}_Q^{(m)}, \mathbf{W}_K^{(m)}, \mathbf{W}_V^{(m)} \in \mathbb{R}^{D_{\text{emb}} \times d_{\text{head}}}\) are the projection matrices, and \(d_{\text{head}}\) is the dimension per head.

The attention output for head \(m\) is:
\begin{equation}
\mathbf{a}^{(l,m)} = \text{softmax}\left( \frac{\mathbf{q}^{(m)} \left(\mathbf{k}^{(m)}\right)^\top}{\sqrt{d_{\text{head}}}} \right) \mathbf{v}^{(m)}.
\end{equation}
The full MHA output is obtained by concatenating all heads:
\begin{equation}
\text{MHA}(\mathbf{z}^{(l)}) = \text{Concat}\left[\mathbf{a}^{(l,1)}, \dots, \mathbf{a}^{(l,h^{(l)})}\right] \mathbf{W}_O,
\end{equation}
where \(\mathbf{W}_O \in \mathbb{R}^{h^{(l)} d_{\text{head}} \times D_{\text{emb}}}\) is the output projection matrix.

The importance score of each attention head \(m\) is computed as:
\begin{equation}
F^{(m)}_{\text{head}} = \sum_{\mathbf{s} \in \mathcal{X}} \left\| \mathbf{a}^{(l,m)}(\mathbf{s}) \right\|_1,
\end{equation}
where \(\mathbf{a}^{(l,m)}(\mathbf{s})\) denotes the output of head \(m\) for input \(\mathbf{s}\).

\vspace{0.5em}
In practice, we use 1024 samples to compute the importance scores. After obtaining all scores, the embedding dimensions, MLP hidden neurons, and MHA heads are individually rearranged in descending order according to their importance.

\begin{figure*}[t]
\centering
\includegraphics[width=1.98\columnwidth]{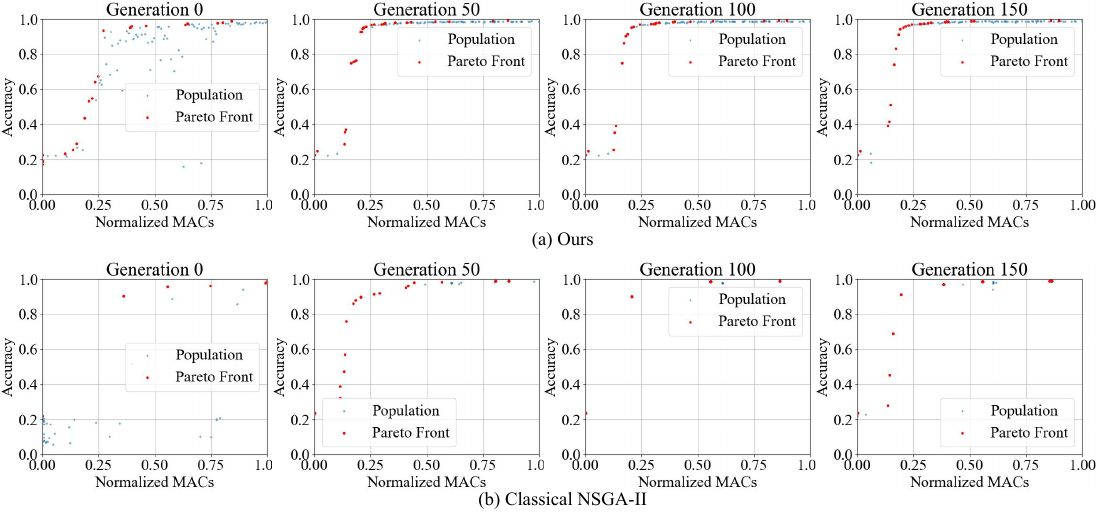} 
\caption{\textbf{Comparison between the proposed customized NSGA-II algorithm (a) and the standard NSGA-II baseline (b).} The customized variant yields a more diverse and well-distributed population across MACs, offering broader coverage and avoiding the redundancy typically exhibited by the standard approach.}
\label{fig:nsga comparision}
\end{figure*}

\subsection{Details of NSGA-II for Submodel Search}
\label{nsga}
In the main text, we briefly introduced how we apply NSGA-II to identify the Pareto front of submodels balancing complexity and accuracy. Here, we provide a detailed description.

\paragraph{Principle.}
NSGA-II is a classic multi-objective evolutionary algorithm designed to find Pareto-optimal solutions. It maintains a population of candidates, evolving them through crossover and mutation operations. Selection is based on fast non-dominated sorting, which ranks solutions into Pareto fronts, and crowding distance, which maintains diversity within the population.

\paragraph{Encoding.}
We encode each submodel configuration as a compact binary sequence.

For the embedding dimension, we reduce the search space by setting \(D_{\text{emb}} = k \times d_{\text{head}}\), where \(k\) is a scaling factor controlling the embedding size and \(d_{\text{head}}\) is the dimension per attention head. The value of \(k\) is encoded using \(k\) binary bits, and during decoding, the final embedding dimension is obtained by summing the active bits and multiplying by \(d_{\text{head}}\).

For the MLP expansion ratio \(r\), we adopt an 8-bit binary encoding, allowing \(r\) to vary between 0.5 and 4.0. Similar to the embedding dimension, the final expansion ratio is derived by summing the activated bits.

For the number of attention heads \(h\) in the MHA module, we again allocate \(k\) binary bits. The number of active heads is determined by summing the corresponding bits during decoding.

Finally, for depth control, we allocate \(L\) binary bits for the MHA path and another \(L\) bits for the MLP path, where \(L\) is the total number of blocks. Each bit specifies whether the corresponding block is retained (\(1\)) or skipped (\(0\)) during submodel instantiation.

\paragraph{Search Procedure.}
During the search phase, we use a mini-batch of 1024 samples randomly drawn from the training set to evaluate candidate submodels. 
However, due to the small batch size and the risk of overfitting on the training set, directly applying standard NSGA-II—based on non-dominated sorting with crowding distance—may result in limited solution diversity and poor coverage of the search space.

To mitigate this, we introduce a partitioned selection strategy. 
Specifically, we divide the normalized MACs range into 20 intervals and perform independent selection within each interval, ensuring that the overall population achieves broad coverage across different model complexity levels.

Within each partition, a minimum complexity difference between selected individuals is enforced to enhance diversity. 
During selection, individuals from the current Pareto front are prioritized. 
For subsequent fronts, crowding distance is recalculated jointly over the candidate set and the already selected individuals. 
Selection is then performed based on the updated crowding distance ranking, encouraging a more diverse and well-distributed submodel population across the entire search space.

\paragraph{Initialization.}
At the initialization stage, we construct the population using only two types of individuals: one with all-zero encoding and one with all-one encoding, corresponding to the two extremes of model complexity. 
New offspring are subsequently generated through crossover and mutation operations. 
Compared to random initialization, this strategy provides better coverage of the search space boundaries and significantly accelerates convergence during early generations.

\paragraph{Setting.}
We set the population size to 100, the crossover probability to 0.95, the mutation probability to 0.3, and the number of iterations to 300.

\paragraph{Results.}
As shown in Figure \ref{fig:nsga comparision}, we compare the performance of the classical NSGA-II and our improved NSGA-II on the SVHN dataset after 0, 50, 100, and 150 iterations, under identical initialization conditions.
We visualize both the population distribution and the corresponding Pareto fronts at each stage.
Our improved NSGA-II exhibits faster convergence, higher-quality solutions, and greater population diversity.
In contrast, the standard NSGA-II suffers from noticeable redundancy, where the population collapses into a few repetitive solutions as the iteration progresses.

\begin{figure*}[t]
\centering
\includegraphics[width=1.98\columnwidth]{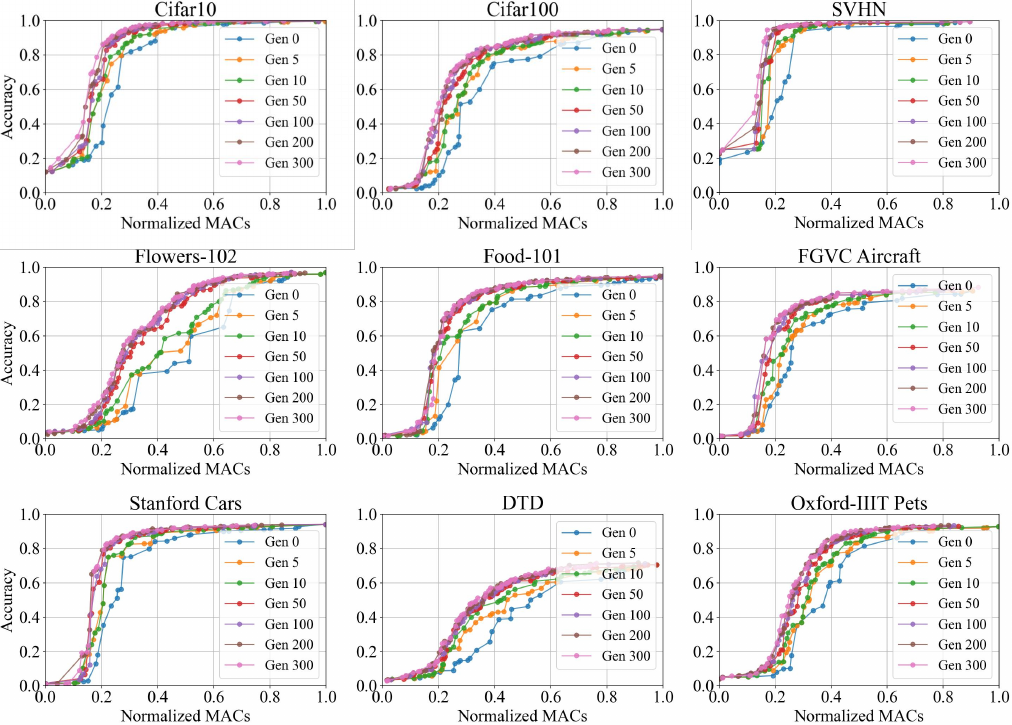} 
\caption{\textbf{Evolution of the Pareto front over generations on nine image classification datasets.}}
\label{fig:nsga presentation}
\end{figure*}

The Figure \ref{fig:nsga presentation} shows the evolution of the Pareto front over generations across nine datasets, demonstrating the effectiveness of our search strategy in progressively identifying high-quality submodels under varying resource constraints.

\section{Training Hyperparameter Settings}
\label{Training Hyperparameter Settings}
\subsection{Curriculum Elastic Adaptation Training Setting}
We conduct our experiments using ViT-Base as the backbone. The training configuration for the Curriculum Elastic Adaptation stage is detailed below.

The model is trained for a total of 100 epochs, with elasticity expansion scheduled at epochs $\{10, 15, 20, 25, 30, 35\}$. At each expansion step, the sampling ranges of the submodel hyperparameters are progressively broadened to increase the model's elasticity.

Initially, the upper bounds are set as:
\[
R_{\max} = 4, \quad H_{\max} = 12, \quad E_{\max} = 768, \quad n_{\max} = 0,
\]
where $R_{\max}$ denotes the maximum MLP expansion ratio, $H_{\max}$ maximum number of attention heads, $E_{\max}$ the maximum embedding dimension, and $n_{\max}$ controls the maximum number of layers that can be skipped.

The decrements applied at each expansion step are configured as:
\[
\Delta_R = \{1, 0.5, 0\}, \quad \Delta_H = 1, \quad \Delta_E = 64, \quad \Delta_n = 1,
\]
where $\Delta_n$ is applied only during the first two expansion steps to gradually introduce layer skipping.

By the end of the Curriculum Elastic Adaptation stage, the sampling ranges are expanded to:
\[
R \in [0.5, 4], \quad H \in [6, 12], \quad E \in [384, 768].
\]

\subsection{Learning Rate and Training Epochs}

For all nine image classification datasets, we trained the model for 100 epochs in both Stage 1 and Stage 2. Except for FGVC~\cite{fgvc} and Stanford Cars~\cite{car}, we used a learning rate that decays from 1e-5 to 1e-7 with linear warm-up and cosine decay in both stages. For FGVC~\cite{fgvc} and Stanford Cars~\cite{car}, we adopted a higher initial learning rate: 1e-4 to 1e-6 in Stage 1, and 1e-5 to 1e-7 in Stage 2.

On segmentation benchmarks ADE20K~\cite{ade20k}, we trained for 25 epochs per stage using a learning rate schedule from 1e-5 to 1e-7 throughout.

For more difficult datasets Imagenet-1K~\cite{imagenet}, Kvasir~\cite{kvasir} and UCMerced~\cite{ucm}, we also followed a 100+100 epoch training schedule, using a learning rate from 1e-5 to 1e-7 in both stages.

All learning rates mentioned above refer to the ViT backbone. The learning rate for the router was set to 1,000 times that of the backbone.

For all baselines, we ensured that the total number of training epochs was kept consistent with our method.

\section{More Results}
\label{More Detailed Results}

\subsection{Comparision wtih other metrics}
\begin{figure}[t]
\centering
\includegraphics[width=1.0\columnwidth]{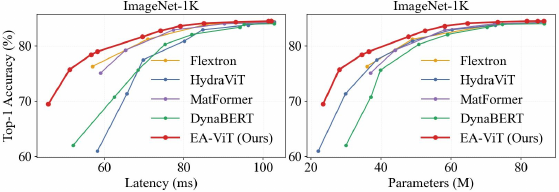} 
\vspace{-0.1in}
\caption{\textbf{Comparison of Top-1 accuracy under different latency and parameter count constraints on the ImageNet-1K dataset.} Our method consistently outperforms existing approaches, demonstrating superior trade-offs not only in computation (MACs), but also in latency and model size, which are critical for real-world deployment.}
\vspace{-0.2in}
\label{fig:metrics}
\end{figure}

While the main text focuses on the MACs–accuracy trade-off, we further assess our method using additional metrics, including latency and parameter count. As shown in Figure~\ref{fig:metrics}, we compare our model with existing approaches on the ImageNet-1K dataset. The results show that our method consistently achieves higher accuracy under varying latency and parameter constraints. This demonstrates that our approach not only achieves a superior trade-off in terms of computational cost, but also performs competitively across other key deployment-related metrics.

\subsection{Comparision under more MACs}
As shown in Table \ref{tab:five_macs_nine_ds}, we compare our method with previous approaches under five MACs levels: 5, 8, 11, 14, and 17 GMACs. Our method consistently achieves superior performance across most settings, with particularly pronounced advantages under lower computational budgets.

\begin{table*}[htbp]
  \centering
  \scriptsize
  \setlength{\tabcolsep}{3pt}
  \renewcommand{\arraystretch}{1.05}

    \caption{\textbf{Comparison of Top-1 accuracy across nine classification benchmarks under five GMACs constraints.} The best results are highlighted in \textbf{bold}, and the second-best are \underline{underlined}.}
  \label{tab:five_macs_nine_ds}

  \resizebox{\textwidth}{!}{%
    \begin{tabular}{@{} l *{9}{c} @{}}
      \toprule
      \multirow{2}{*}{Method} &
        \multicolumn{9}{c}{Dataset} \\
      \cmidrule(lr){2-10}
        & Cifar 10\cite{cifar100} & Cifar 100 \cite{cifar100} & SVHN\cite{svhn} & Flowers \cite{flower102} & Food101 \cite{food101} & Aircraft \cite{fgvc} & Cars \cite{car} & DTD \cite{DTD} & Pets \cite{pet} \\ \midrule

      \multicolumn{10}{c}{\textbf{5 GMACs Budget}} \\ \midrule
      DynaBERT\cite{dynabert}   & -- & -- & -- & -- & -- & -- & -- & -- & -- \\
      MatFormer\cite{matformer} & -- & -- & -- & -- & -- & -- & -- & -- & -- \\
      HydraViT\cite{hydravit}   & \underline{91.99} & \underline{73.26} & \underline{96.26} & \underline{12.09} & \underline{73.92} & \underline{70.66} & \underline{78.98} & \underline{21.31} & \underline{31.36} \\
      Flextron\cite{flextron}   & -- & -- & -- & -- & -- & -- & -- & -- & -- \\
      \rowcolor{gray!10}
      EA-ViT (ours)             & \textbf{95.81} & \textbf{80.49} & \textbf{97.49} & \textbf{64.85} & \textbf{83.32} & \textbf{79.06} & \textbf{86.91} & \textbf{59.99} & \textbf{78.95} \\[2pt]

      \multicolumn{10}{c}{\textbf{8 GMACs Budget}} \\ \midrule
      DynaBERT\cite{dynabert}   & 94.24 & 78.30 & 96.58 & 24.53 & 82.24 & 71.19 & 80.19 & 36.88 & 57.05 \\
      MatFormer\cite{matformer} & 96.17 & \underline{86.18} & 97.45 & 69.00 & 86.31 & 76.13 & 88.10 & 49.25 & 84.53 \\
      HydraViT\cite{hydravit}   & 96.11 & 85.39 & 96.98 & 29.27 & 85.49 & 75.95 & 85.07 & 43.53 & 75.32 \\
      Flextron\cite{flextron}   & \underline{97.11} & 85.95 & \underline{97.61} & \underline{73.80} & \underline{87.79} & \underline{79.36} & \underline{88.62} & \underline{61.21} & \underline{86.35} \\
      \rowcolor{gray!10}
      EA-ViT (ours)            & \textbf{97.98} & \textbf{88.20} & \textbf{97.63} & \textbf{85.39} & \textbf{90.05} & \textbf{83.37} & \textbf{90.09} & \textbf{67.56} & \textbf{90.57} \\[2pt]

      \multicolumn{10}{c}{\textbf{11 GMACs Budget}} \\ \midrule
      DynaBERT\cite{dynabert}   & 97.53 & 88.29 & 96.96 & 62.46 & 88.61 & 74.61 & 83.47 & 58.18 & 83.04 \\
      MatFormer\cite{matformer} & 98.08 & 90.29 & 97.65 & \underline{88.00} & 90.52 & 78.93 & \underline{90.49} & 64.86 & 91.64 \\
      HydraViT\cite{hydravit}   & 98.08 & 89.89 & 97.24 & 70.46 & 89.58 & 77.94 & 87.18 & 63.38 & 89.12 \\
      Flextron\cite{flextron}   & \underline{98.38} & \underline{90.34} & \textbf{97.72} & 82.51 & \underline{90.85} & \underline{79.86} & 90.24 & \underline{69.54} & \textbf{92.85} \\
      \rowcolor{gray!10}
      EA-ViT (ours)            & \textbf{98.53} & \textbf{90.64} & \underline{97.70} & \textbf{93.45} & \textbf{91.20} & \textbf{84.78} & \textbf{91.28} & \textbf{71.97} & \underline{92.49} \\[2pt]

      \multicolumn{10}{c}{\textbf{14 GMACs Budget}} \\ \midrule
      DynaBERT\cite{dynabert}   & 98.17 & 91.07 & 97.42 & 75.70 & 90.85 & 75.93 & 87.78 & 65.70 & 89.85 \\
      MatFormer\cite{matformer} & \underline{98.76} & \underline{92.14} & \underline{97.77} & \underline{94.61} & \underline{92.21} & \underline{80.11} &\underline{91.37} & 71.86 & 93.37 \\
      HydraViT\cite{hydravit}   & 98.64 & 91.98 & 97.62 & 85.46 & 91.67 & 79.05 & 89.19 & 69.81 & 92.29 \\
      Flextron\cite{flextron}   & 98.71 & 91.88 & 97.71 & 84.98 & 91.94 & 80.05 & 91.09 & \underline{72.68} & \textbf{93.87} \\
      \rowcolor{gray!10}
      EA-ViT (ours)            & \textbf{98.96} & \textbf{92.62} & \textbf{97.81} & \textbf{96.13} & \textbf{92.73} & \textbf{85.47} & \textbf{91.95} & \textbf{74.94} & \underline{93.55} \\[2pt]

      \multicolumn{10}{c}{\textbf{17 GMACs Budget}} \\ \midrule
      DynaBERT\cite{dynabert}   & 98.89 & 92.95 & 97.76 & 83.05 & 92.77 & 79.53 & 90.77 & 73.05 & 93.53 \\
      MatFormer\cite{matformer} & \underline{99.03} & 92.91 & \textbf{97.87} & \textbf{96.96} & \underline{93.07} & \underline{80.84} & \underline{91.76} & \textbf{75.29} & 94.06 \\
      HydraViT\cite{hydravit}   & 98.91 & \underline{93.13} & 97.83 & 90.75 & 92.98 & 79.21 & 89.99 & 73.72 & 93.68 \\
      Flextron\cite{flextron}   & 98.79 & 92.61 & 97.77 & 85.97 & 93.02 & 80.52 & 91.23 & 73.56 & \underline{94.16} \\
      \rowcolor{gray!10}
      EA-ViT (ours)            & \textbf{99.09} & \textbf{93.22} & \underline{97.84} & \underline{95.79} & \textbf{93.13} & \textbf{85.48} & \textbf{92.17} & \underline{74.93} & \textbf{94.32} \\

      \bottomrule
    \end{tabular}%
  }
\end{table*}

\begin{figure*}[t]
\centering
\includegraphics[width=1.98\columnwidth]{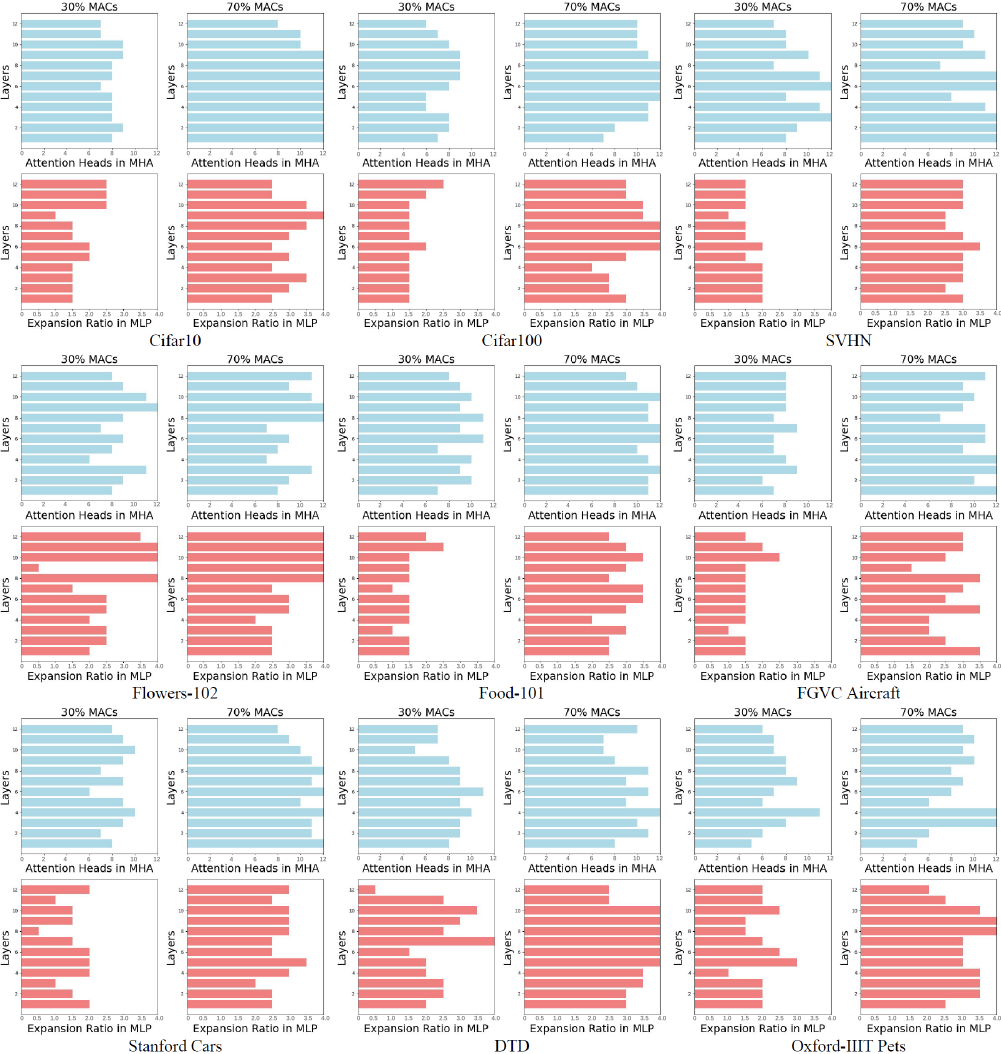} 
\caption{\textbf{Layer-wise visualization of submodel architectures under normalized MACs constraints of 0.3 and 0.7 across nine datasets.} For each submodel, we plot the MLP expansion ratio and the number of attention heads in each layer. The comparison reveals distinct architectural patterns that emerge under different computational budgets and dataset characteristics.}
\label{fig：suppl presentation1}
\end{figure*}

\begin{figure*}[t]
\centering
\includegraphics[width=1.98\columnwidth]{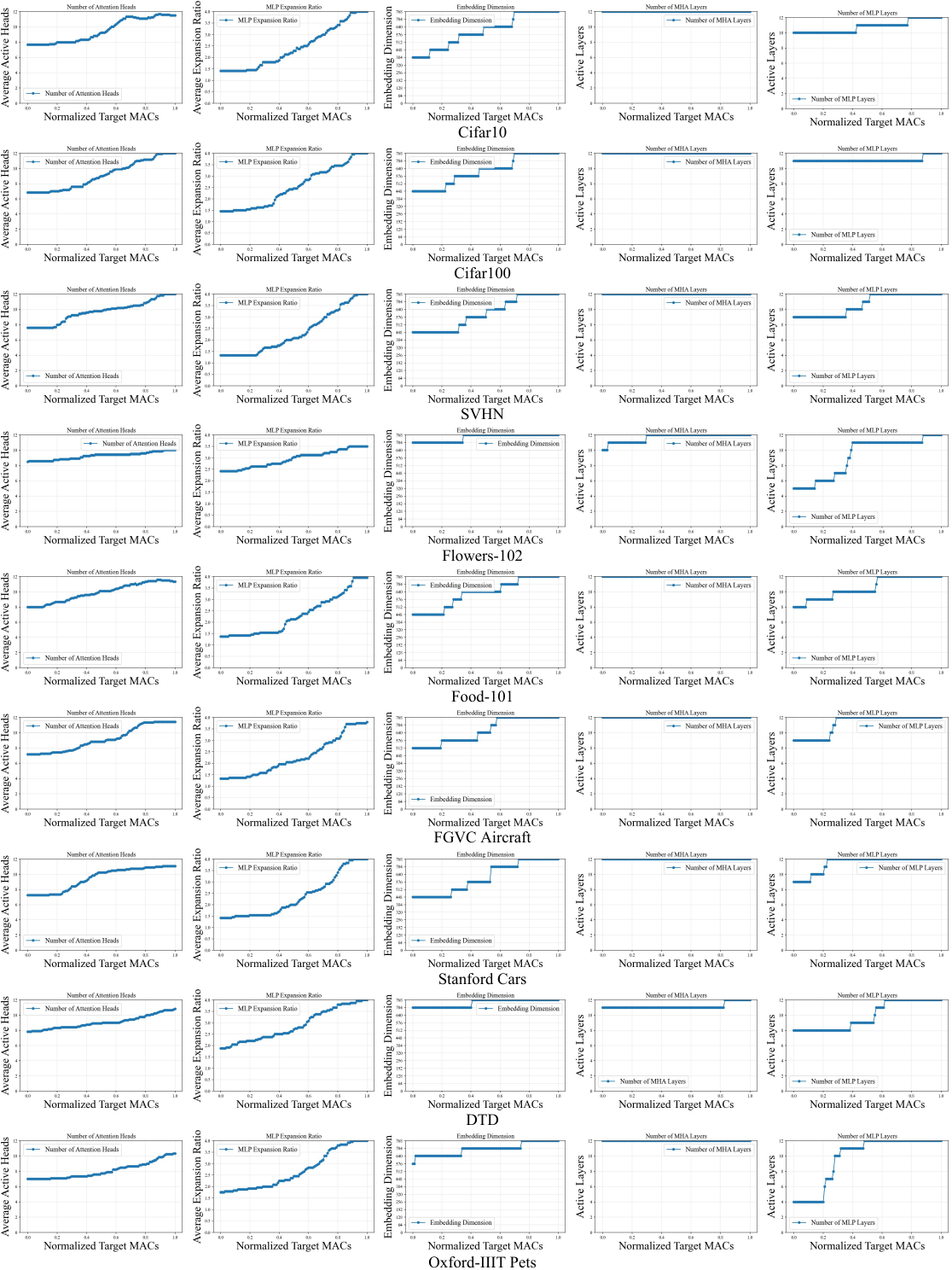} 
\caption{\textbf{Trends in architectural configurations selected by the router as the normalized MACs constraint varies.} For each of the nine datasets, we report the average number of MHA attention heads, MLP expansion ratio, embedding dimension, and the number of MHA and MLP layers. The results demonstrate how the router dynamically adapts submodel complexity based on resource constraints and dataset demands.}
\label{fig:suppl presentation2}
\end{figure*}

\section{Visualization of Submodel Architectures}
\label{Visualization of Submodel Architectures}
As shown in Figure \ref{fig：suppl presentation1}, we visualize the architectural configurations of submodels on nine datasets under normalized MACs constraints of 0.3 and 0.7. Specifically, we present the per-layer MLP expansion ratios and the number of attention heads in the MHA modules, revealing clear differences across datasets and computational budgets.

In addition, we illustrate how the router's output evolves as the normalized MACs constraint varies in Figure \ref{fig:suppl presentation2}. For each dataset, we report the average number of attention heads, MLP expansion ratios, embedding dimensions, and the number of MHA and MLP layers selected by the router. These results provide insight into how the model adjusts architectural complexity in response to resource constraints and dataset characteristics.

\section{Potential Applications}
\label{Potential Applications}
Although this work focuses on Vision Transformers and primarily targets image classification and segmentation tasks, the proposed EA-ViT framework can be seamlessly applied to other Transformer-based architectures and extended to broader domains such as image and video editing~\cite{ma2022visual, yan2025eedit, ma2025followcreation}. Furthermore, incorporating more deployment-efficient datasets~\cite{yang2024clip, yang2024adaaugment, yang2024investigating} may further enhance both efficiency and performance in real-world scenarios.

\end{document}